\documentclass{midl} 
\usepackage{mwe} % to get dummy images
\usepackage{placeins}
\jmlrvolume{-- nnn}
\jmlryear{2026}
\jmlrworkshop{Full Paper -- MIDL 2026}
\editors{Accepted for publication at MIDL 2026}

\usepackage{multirow}
\usepackage{xcolor}
\usepackage{soul}

\definecolor{lightyellow}{RGB}{255, 249, 196}
\sethlcolor{lightyellow}

\title[Multimodal Deep Learning for Pancreatic Cancer Resectability]{Multimodal Assessment of Pancreatic Cancer Resectability Using Deep Learning}

\midlauthor{
\Name{Vincent Ochs\nametag{$^{1}$}} \orcid{0000-0003-0237-7496} \Email{vincent.ochs@unibas.ch}\\
\Name{Christoph Kuemmerli\nametag{$^{2}$}} \Email{Christoph.Kuemmerli@clarunis.ch}\\
\Name{Florentin Bieder\nametag{$^{1}$}} \Email{florentin.bieder@unibas.ch}\\
\Name{Julia Wolleb\nametag{$^{1}$}} \Email{julia.wolleb@unibas.ch}\\
\Name{Jo\"el L.~Lavanchy\nametag{$^{1,2}$}} \orcid{0000-0003-0248-4996} 
\Email{joel.lavanchy@clarunis.ch}\\
\Name{Julia Ruppel\nametag{$^{2}$}} \Email{Julia.Ruppel@clarunis.ch}\\
\Name{Jan Liechti\nametag{$^{2}$}} \Email{j.liechti@unibas.ch}\\
\Name{Stephanie Taha-Mehlitz\nametag{$^{2}$}} \Email{stephanie.taha@clarunis.ch}\\
\Name{Christian Andreas Nebiker\nametag{$^{3}$}} \Email{Christian.Nebiker@ksa.ch}\\
\Name{Beat M\"uller\nametag{$^{2}$}} \Email{Beat.Mueller@clarunis.ch}\\
\Name{Giuseppe Kito Fusai\nametag{$^{4}$}} \Email{G.fusai@nhs.net}\\
\Name{Joerg-Matthias Pollok\nametag{$^{4}$}} \Email{joerg-matthias.pollok@nhs.net}\\
\Name{Anas Taha\nametag{$^{1,5,6}$}} \Email{anas.taha@unibas.ch}\\
\Name{Philippe C.~Cattin\midljointauthortext{Shared last authorship}\nametag{$^{1}$}}  \orcid{0000-0001-8785-2713}\Email{philippe.cattin@unibas.ch}\\
\Name{Sebastian Staubli\midljointauthortext{Shared last authorship}\nametag{$^{2,4}$}} \Email{s.staubli@nhs.net}\\[0.5em]
\addr $^{1}$ University of Basel, Department of Biomedical Engineering, 4123 Allschwil, Switzerland \\
\addr $^{2}$ Clarunis, University Digestive Health Centre, Basel, Switzerland \\
\addr $^{3}$ Department of General Surgery, Kantonsspital Aarau, Tellstrasse, 5000, Aarau, Switzerland \\
\addr $^{4}$ HPB and Liver Transplant Service, Royal Free Hospital, London, United Kingdom \\
\addr $^{5}$ Department of Visceral Surgery, Cantonal Hospital Basel-Land, Liestal, Switzerland \\
\addr $^{6}$ Department of Surgery, East Carolina University, Brody School of Medicine, Greenville, North Carolina USA
}

\begin{document}

\maketitle

\begin{abstract}
Accurate determination of pancreatic ductal adenocarcinoma (PDAC) resectability relies on evaluating how the tumor interacts with major peripancreatic vessels on CT imaging, yet expert assessment often shows substantial variability. We introduce a fully automated multimodal deep learning framework that jointly analyzes 3D contrast enhanced CT and structured clinical information to classify patients into the three National Comprehensive Cancer Network (NCCN) resectability categories (upfront resectable, borderline resectable, locally advanced). The approach uses a Swin-UNETR backbone to obtain anatomy aware image representations through auxiliary segmentation of pancreas, tumor, and vascular structures. These features are fused with a compact clinical embedding derived from 17 routinely collected variables and processed by a lightweight classification head. Model training is guided by a dynamic multitask objective that adapts the balance between segmentation and classification based on current tumor Dice performance, promoting feature representations that remain both anatomically informed and discriminative.

In a cohort of 159 patients (85 upfront resectable, 47 borderline resectable, 27 locally advanced), the proposed method achieved an AUC of 0.86, a macro-F1 of 0.79, and an accuracy of 0.85 using stratified nested 5-fold cross validation, outperforming adapted transformer based and geometric baseline approaches. External validation on an independent cohort with 52 patients from Kantonsspital Aarau (KSA Aarau) yielded an AUC of 0.86, a macro-F1 of 0.81, and an accuracy of 0.87, supporting cross-institution generalization.

Notably, the external KSA Aarau cohort contained complete clinical information for all variables used by the model and therefore did not require imputation. The comparable performance observed on this dataset suggests that the KNN based imputation applied to the training cohort did not introduce a detectable performance bias for the clinical variables considered.

 Because segmentation labels are required only during training, the final system enables mask free inference while preserving vessel aware interpretability. These findings demonstrate that integrating anatomical supervision with clinical context yields a robust and reproducible tool for supporting operability (i.e., NCCN-based resectability) assessment in pancreatic cancer.
The implementation is publicly available at 
\url{https://github.com/vincentochs/pancreas_resectability},
and the data, as well as the weights, can be made available by the corresponding author upon reasonable request.
\end{abstract}

\begin{keywords}
PDAC, Multi-Modality Model, Adaptive Loss Schedule, Resectability Classification.
\end{keywords}

\section{Introduction}
Pancreatic ductal adenocarcinoma (PDAC) is among the most lethal malignancies, accounting for over 90\% of pancreatic cancers and exhibiting a 5 year survival rate below 10\% \citep{Sung2021,Siegel2023}. Owing to its aggressive growth and the absence of specific early symptoms, pancreatic cancer is often diagnosed only at a late stage \citep{Rawla2019}. Complete surgical resection (R0) remains the only curative option, yet fewer than 20\% of patients are eligible for surgery \citep{Tempero2021,Neoptolemos2018}.

Resectability is primarily determined by tumor involvement of key peripancreatic vessels, including the superior mesenteric artery (SMA), superior mesenteric vein (SMV), portal vein, and celiac trunk \citep{Katz2013,Bockhorn2014}. The NCCN defines three resectability categories, namely upfront resectable, borderline resectable, and locally advanced, based on these vascular relationships \citep{Tempero2021}. Accurate assessment on contrast enhanced computed tomography (CT) is critical for treatment planning \citep{Isaji2018,Callery2009}, yet inter-observer agreement remains poor, often below 70\% even under standardized criteria \citep{Giannone2021,Katz2013}. As a result, operability assessment frequently relies on qualitative judgment rather than reproducible quantitative measurements. In this study, operability refers specifically to NCCN-based resectability categories, i.e., upfront resectable, borderline resectable, and locally advanced disease \citep{Tempero2021}.

Deep learning offers a path toward standardized CT interpretation by automatically extracting spatial and textural features \citep{Litjens2017,Li}. However, most prior works focus on segmentation \citep{Lim,Zhou2020} or unimodal imaging based prediction \citep{Huang}, without integrating clinical variables such as comorbidities, tumor markers (CA19-9, CEA), and treatment history that influence decisions in multidisciplinary tumor boards \citep{Hartwig2013,Rahib2014}. Many segmentation-based pipelines require segmentation masks at inference (manual or from an additional segmentation step), which can complicate clinical deployment and workflow integration \citep{Bereska2024,Viviers2023}.

To address these limitations, we propose an end-to-end multimodal framework that integrates 3D CT imaging with structured clinical data to predict NCCN resectability. By leveraging auxiliary anatomical supervision during training, the model learns vessel aware representations that generalize to unseen cases without requiring segmentation masks at test time, enabling a fully automated, interpretable decision support system for NCCN based resectability.

\paragraph{Contributions.}
The main contributions of this work are: 
(i) a clinically motivated multimodal framework integrating 3D CT imaging with structured clinical variables for NCCN resectability prediction; 
(ii) anatomy guided auxiliary supervision that enforces vessel aware feature learning without requiring segmentation masks at inference; 
(iii) a performance adaptive multitask objective that dynamically balances segmentation and classification throughout training; and 
(iv) a comprehensive evaluation including ablations and comparisons demonstrating improved predictive performance and clinical consistency over existing approaches.

\section{Related Work} 
Automated resectability assessment in pancreatic cancer has largely focused on CT-based quantification of tumor vessel relationships. Bereska et al.\ used segmentation derived tumor–vessel contact lengths to approximate NCCN classes \citep{Bereska2024}, while Viviers et al.\ combined multiorgan and vessel segmentation with handcrafted geometric features for resectability prediction \citep{Viviers2023}. 

Although effective, these pipelines require accurate segmentations at inference and do not incorporate patient level clinical information such as comorbidities or tumor markers. More recent approaches explore direct CT-based prediction of resectability or prognosis, but remain unimodal and often lack alignment with NCCN surgical frameworks \citep{Schuurmans2025_EndToEndPDAC}. 
For anatomical representation learning, convolutional encoder–decoders such as U-Net and its 3D variants are standard for abdominal organ and vessel segmentation \citep{Ronneberger2015}. Transformer based models like SwinUNETR further improve fine vascular delineation via shifted window self attention \citep{cao2021swinunetunetlikepuretransformer}. 

Pretraining on large abdominal datasets (e.g., BTCV, AMOS) enhances small structure generalization and accelerates downstream optimization \citep{ji2022amoslargescaleabdominalmultiorgan}. Multitask learning has also been explored, though most works rely on fixed loss-weighting between tasks and primarily emphasize anatomical accuracy rather than clinically actionable prediction \citep{KORDNOORI2024101931}.

The integration of imaging with clinical variables is gaining traction in oncology. Early multimodal approaches fused radiomic or handcrafted imaging features with tabular data via late concatenation, offering limited cross-modal interaction \citep{jiang2021fusionmedicalimagingelectronic,Huang2016}. 
More recent transformer-based models enable richer multimodal reasoning. For example, the Texture-Aware Transformer (TAT) captures texture variations across multi-phase CT acquisitions for PDAC prognosis \citep{dong2023improvedprognosticpredictionpancreatic}, while the Transformer-based Multimodal Network for Segmentation and Survival prediction (TMSS) fuses CT, PET, and clinical data for head-and-neck cancer survival prediction \citep{saeed2022tmssendtoendtransformerbasedmultimodal}.

However, these works focus on longterm outcomes, not immediate surgical resectability, and generally omit explicit anatomical supervision. The present study builds on these foundations by combining anatomy guided representation learning with structured clinical fusion to deliver a fully automated NCCN-based resectability prediction model that is trained using NCCN based labels and anatomy guided supervision, without requiring manual segmentations at inference.

\section{Method}
\begin{figure}
\includegraphics[width=\textwidth]{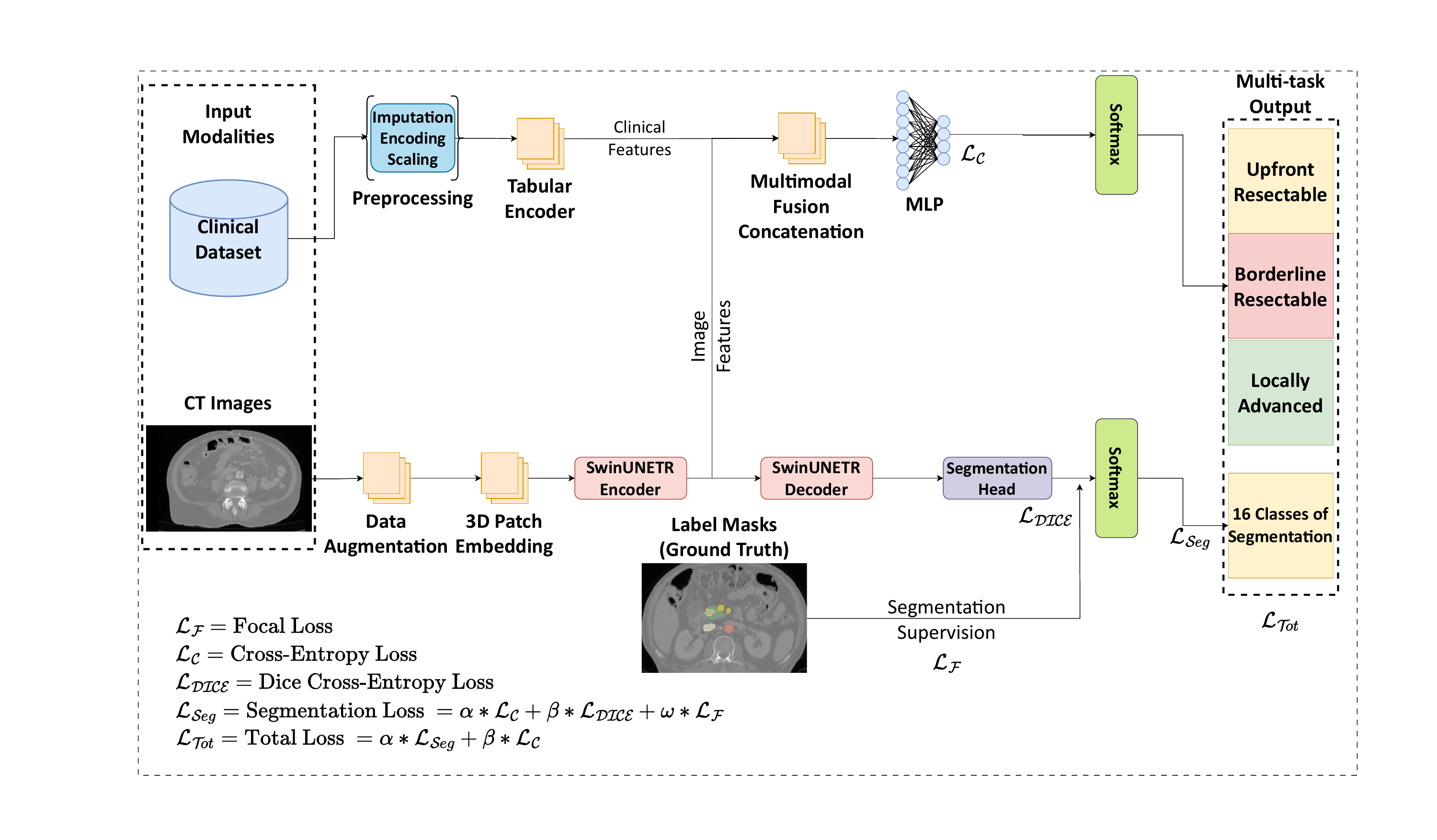}
\caption{Proposed Network Architecture.
During backpropagation, gradients from both the segmentation decoder and the classification fusion head are propagated through the shared encoder. 
This joint optimization enables the encoder to learn anatomically meaningful, discriminative features for resectability prediction.
}
\label{fig:Fig1}
\end{figure}

We propose an end-to-end multimodal framework that predicts NCCN resectability from contrast enhanced CT and structured clinical data (Fig.~\ref{fig:Fig1}). 
The model combines a 3D anatomy aware Swin-UNETR encoder–decoder with a tabular fusion head, trained jointly under a dynamic multitask objective. From each CT volume, the encoder produces a 256-dimensional anatomical feature vector while the decoder performs auxiliary multiorgan segmentation. In parallel, 17 clinical variables are processed through a multilayer perceptron (MLP) to yield a 32-dimensional clinical embedding. Both embeddings are concatenated into a 288-dimensional representation and passed to a classification head to predict the three NCCN resectability classes. During training, segmentation and classification gradients update the shared encoder, encouraging anatomically grounded and task relevant representations.

\subsection{Data Modalities and Preprocessing}

\paragraph{CT Imaging:} 
Each volumetric CT scan is provided in NIfTI format and resampled to isotropic 1\,mm spacing before being cropped or padded to $160 \times 160 \times 160$ voxels. This resolution preserves peripancreatic vessel continuity while keeping GPU memory usage within feasible limits. Voxel intensities are z-score normalized to account for acquisition variability. To improve robustness and generalization across scanners, we apply MONAI based augmentations including random flips, 90° rotations, affine transformations (rotation range $\pm 0.1$~rad), Gaussian noise ($\sigma=0.01$), and random intensity shifts ($\pm 0.1$) \citep{cardoso2022monaiopensourceframeworkdeep}.

The CT data were acquired at three institutions. To mitigate inter-site variability, all volumes were resampled to a common isotropic resolution, intensity-normalized, and processed using an identical preprocessing and augmentation pipeline. Clinical variables were harmonized using consistent definitions and units across institutions prior to model training.

\paragraph{Segmentation Masks:}
Voxelwise annotations were created and quality-controlled in a multi-expert workflow by up to five physicians.
They comprise 16 anatomical and pathological classes: background, pancreas, tumor, portal vein, splenic vein, superior mesenteric vein (SMV), superior mesenteric artery (SMA), celiac trunk, aorta, inferior vena cava, common hepatic artery, proper hepatic artery, gastroduodenal artery, splenic artery, dilated pancreatic duct, and bile duct. The segmentation loss inherently emphasizes tumor and major vessels (SMA, SMV, portal vein, celiac trunk) because the Dice and Focal components give higher relative importance to small, clinically relevant structures. 

Because the segmentation decoder and classification head share a common encoder, 
this emphasis encourages the encoder to learn vessel-aware anatomical features that 
are subsequently used for multimodal fusion and resectability classification.
All masks were resampled to isotropic 1\,mm spacing, cropped or padded to the $160 \times 160 \times 160$ field-of-view used for CT preprocessing, and aligned with the normalized CT volumes. These preprocessed masks serve as ground truth for auxiliary segmentation supervision during training.

\paragraph{Clinical Data:}
Seventeen structured variables capture complementary prognostic information, encompassing age at operation or diagnosis, sex, body mass index (BMI), Charlson Comorbidity Index (CCI), American Society of Anesthesiologists (ASA) score, diabetes status, nicotine use, alcohol consumption, CA~19-9 serum baseline level, CA~19-9 delta, CA~19-9 serum level after neoadjuvant therapy, CEA serum level, blood glucose, HbA1c, bilirubin at initial diagnosis, histopathological grade (G), and neoadjuvant (radio-)chemotherapy status.

The selected clinical variables were pre-defined in collaboration with experienced clinicians and reflect routinely available factors considered during multidisciplinary tumor board decision making rather than an exhaustive set of all disease-associated variables.
Overall missingness across the 17 clinical variables was low (less than 5\% of entries), with most variables being complete or near-complete. The highest missingness was observed for HbA1c (5\%).
Missing values were imputed hierarchically: KNN ($k=5$) for continuous variables and mode imputation for categorical attributes. 
Categorical variables were one-hot encoded, and continuous features were z-score standardized 
(i.e., transformed to zero mean and unit variance based on the training split).

The processed inputs are passed through a three layer MLP (hidden dimensions: 64 and 32, each followed by LayerNorm and ReLU), producing a final 32-dimensional clinical embedding.

\subsection{Model Architecture}

\paragraph{Swin-UNETR Backbone:}
We adopted the Swin-UNETR \citep{cao2021swinunetunetlikepuretransformer} as the imaging encoder–decoder backbone. 
Its shifted window self-attention captures both local and global 3D contexts, crucial for fine vascular delineation around the pancreas. 
The encoder and decoder share skip connections for anatomical reconstruction, while the encoder bottleneck outputs a 256-dimensional feature vector used for both segmentation and classification.

This bottleneck representation constitutes a merged embedding of local and global features learned through Swin-UNETR’s hierarchical, shifted-window self-attention.

Pretraining on the BTCV dataset provides anatomical priors and accelerates convergence.

\paragraph{Multimodal Fusion:}
To integrate imaging and clinical information, the 256-dimensional imaging feature is concatenated with the 32-dimensional clinical embedding to form a 288-dimensional multimodal vector. This fused representation is processed by an MLP fusion head (LayerNorm, ReLU, dropout) and a softmax classifier to predict the three NCCN-based resectability categories. Gradients from both segmentation and classification update the shared encoder, enforcing anatomically informed features. We use simple concatenation for fusion, as ablation experiments showed no measurable benefit from more complex cross-attention mechanisms.

\subsection{Multi-Task Learning Objective}

The model was trained via a joint objective combining segmentation and classification tasks:
\begin{equation}
\mathcal{L}_\text{total} = w_\text{seg}(\tau)\,\mathcal{L}_\text{seg} + w_\text{cls}(\tau)\,\mathcal{L}_\text{cls},
\label{eq:total_loss}
\end{equation}
where $\tau$ denotes the tumor Dice score and $w_{\text{seg}}$ and $w_{\text{cls}}$ 
are the segmentation and classification loss weights, respectively. 
Both weights change dynamically as a function of the current segmentation performance 
$\tau$. This adaptive weighting gradually shifts emphasis from segmentation to classification as segmentation accuracy improves:
early in training, when $\tau$ is low, a higher $w_{\text{seg}}$ emphasizes anatomical 
learning, whereas with improving segmentation quality the weight shifts toward 
classification by increasing $w_{\text{cls}}$. In the following, we discuss these 
components in more detail.

\paragraph{Segmentation Loss:}
The segmentation decoder is optimized using a hybrid loss that combines cross-entropy, Dice, and focal components:
\begin{equation}
\mathcal{L}_\text{seg} = \alpha \,\mathcal{L}_\text{CE} + \beta \,\mathcal{L}_\text{Dice} + \omega \,\mathcal{L}_\text{Focal}.
\label{eq:seg_loss}
\end{equation}
This encourages precise boundary delineation and robustness to class imbalance, particularly for small vessels.

\paragraph{Classification Loss:}
To account for the class imbalance in our dataset (85 upfront resectable, 47 borderline resectable, 27 locally advanced cases), 
we apply a class-weighted cross-entropy loss:
\begin{equation}
\mathcal{L}_\text{cls} = -\sum_{k=1}^{3} v_k \, y_k \log \hat{y}_k,
\label{eq:cls_loss}
\end{equation}
where $y = (y_1, y_2, y_3)$ denotes the one-hot encoded ground-truth class label, 
$\hat{y} = (\hat{y}_1, \hat{y}_2, \hat{y}_3)$ are the predicted class probabilities from the softmax output, 
and $v = [1.0,\, 2.0,\, 2.5]$ are smoothed inverse-frequency weights that increase the contribution of 
minority classes (borderline resectable, locally advanced) without overcompensating for their smaller sample sizes. 
These values were selected empirically to balance gradient magnitudes and ensure stable convergence across folds. 
A sensitivity analysis comparing different weighting schemes (e.g., strict inverse frequency and unweighted baselines) 
is presented in the appendix (see Table~\ref{tab:class_weight_ablation}), demonstrating that the proposed configuration 
achieves the best trade-off between classification AUC and stability across runs.

\paragraph{Adaptive Weighting:}
The relative importance of segmentation and classification is adjusted dynamically according to the current segmentation performance:
\begin{equation}
w_\text{seg}(\tau) :=
\begin{cases}
3.0 & \tau \in [0, 0.1) \\
1.5 & \tau \in [0.1, 0.5) \\
1.0 & \tau \in [0.5, 1]
\end{cases}
\quad
w_\text{cls}(\tau) := \frac{1}{w_\text{seg}(\tau)}.
\label{eq:adaptive_weights}
\end{equation}
The thresholds reflect typical stages of network convergence observed during pilot training.
Early epochs often yield tumor Dice scores below 0.1 due to unstable vessel localization, 
mid training stabilizes around 0.3-0.5 as anatomical structures emerge, 
and values above 0.5 indicate sufficiently reliable feature representations. 
The corresponding weights (3.0, 1.5, 1.0) were empirically chosen to maintain stable gradients, strongly emphasizing anatomical supervision when feature learning is unstable, 
then progressively shifting focus toward the classification objective as segmentation accuracy improves. 
This strategy mirrors a self paced or curriculum learning scheme, 
allowing the model to first learn anatomical context before optimizing the downstream NCCN-based resectability prediction. 
A detailed ablation of alternative threshold and weighting configurations, demonstrating the stability and performance benefits of the proposed schedule, 
is provided in the appendix (see Table~\ref{tab:adaptive_ablation}).

\section{Experiments and Results}

\subsection{Dataset}

This retrospective multimodal cohort comprised 159 patients with histologically confirmed PDAC from the University Hospital Basel and St. Clara Hospital Basel. 
All data were de-identified and collected under institutional review board (IRB) approval. 
Ground-truth NCCN resectability labels were defined during routine clinical care by the multidisciplinary tumor board (MDT) at University Hospital Basel and St. Clara Hospital Basel, involving multiple physicians (e.g., surgery, radiology, oncology) according to NCCN criteria. To further ensure label consistency, MDT-based labels (upfront resectable, borderline resectable, locally advanced) were independently reviewed and validated by an additional physician who was not involved in model development.
For external validation, we additionally collected an independent cohort of 52 patients from Kantonsspital Aarau (KSA Aarau), which was not used for model development or internal cross-validation.

According to NCCN criteria, the cohort included 85 upfront resectable (53.5\%), 47 borderline resectable (29.5\%), and 27 locally advanced cases (17\%), 
each with corresponding contrast enhanced CT imaging and structured clinical data.
The external dataset included 28 upfront resectable cases, 14 borderline resectable cases, and 10 locally advanced cases.

\subsection{Experimental Setup}
We employed a stratified nested 5-fold cross-validation scheme to ensure balanced representation of resectability categories and to obtain an unbiased performance estimate.
For each outer fold, 20\% of the dataset was held out as an independent test set and was never used during training, hyperparameter tuning, or early stopping. The remaining 80\% served as the training pool for an inner split, where we partitioned the data into 64\% training and 16\% validation. This inner validation set was used exclusively for model selection, early stopping, and tuning of segmentation and classification losses. Training used the AdamW optimizer with cosine annealing scheduling, an initial learning rate of $10^{-4}$, weight decay of $10^{-5}$, and early stopping after 20 epochs without improvement on the inner validation set. The batch size was fixed at 16. Hyperparameters
were optimized within the inner loop via grid search over learning rate $\{10^{-3},\,5\!\cdot\!10^{-4},\,10^{-4}\}$, feature size $\{24, 32, 48\}$, batch size $\{8, 16\}$, and segmentation loss coefficients $(\alpha,\beta,\omega)\!\in\![0.2,0.6]$. For each outer fold, the configuration yielding the highest inner validation macro-F1 score was selected and subsequently evaluated on the held-out outer test set. Final performance metrics were averaged across all five outer folds.

All models were implemented in PyTorch using the MONAI framework and trained on an NVIDIA A100 (40GB) GPU. During inference, 3D CT volumes were processed using a sliding window strategy with Gaussian blending \citep{cardoso2022monaiopensourceframeworkdeep}. Segmentation outputs were used only for evaluation, whereas the classification branch generated NCCN resectability predictions without requiring segmentation masks at test time.

\subsection{Evaluation Metrics}
Model performance was assessed for both segmentation and classification. 
Segmentation quality was measured using the Dice Similarity Coefficient (DSC), with emphasis on tumor and vessel classes relevant for surgical planning. Classification performance was evaluated using Accuracy, macro averaged F1-score, and the Area Under the ROC Curve (AUC), capturing overall discrimination and classwise balance across the three NCCN resectability categories.

\subsection{NCCN Geometry Analysis}

To assess whether the model implicitly learns NCCN related geometric patterns, we performed a post-hoc analysis of tumor vessel contact angles on the held out outer test sets. For each patient, predicted multiorgan segmentations were used to extract approximate arterial and venous contact angles around key vessels (arterial: SMA or celiac trunk; venous: SMV or portal vein). Although NCCN resectability definitions consider several factors such as encasement, deformity, occlusion, and reconstructability angle based surrogates provide a simplified approximation: arterial involvement of 1-180° often corresponds to borderline resectability, while $\geq 180^\circ$ indicates locally advanced disease; similarly, venous contact $\geq 180^\circ$ may suggest borderline involvement.

These surrogate angles were plotted in a 2D plane and compared with ground truth NCCN classes and model predictions. The observed clusters show that upfront resectable, borderline, and locally advanced cases tend to occupy the expected angles, suggesting that the model’s anatomical representations capture geometric behavior despite the absence of angle-based supervision. This analysis was purely diagnostic and did not influence training, validation, or model selection. The visualization is shown in Figure~\ref{fig:nccn_scatter} in the appendix.

\subsection{Results}
Although segmentation is used only as an auxiliary task, it plays a central role in shaping the anatomy aware encoder. The Swin-UNETR decoder produced stable multiorgan segmentations 
across folds, achieving DSC values of $0.82 \pm 0.03$ for pancreas, 
$0.71 \pm 0.05$ for tumor, and $0.67 \pm 0.06$ for major vessels. Representative qualitative examples illustrating pancreas, tumor, and major vessel 
segmentations are shown in Appendix Figure~\ref{fig:qualitative_segmentation}.
The lower vessel performance reflects the intrinsic difficulty of segmenting small, low contrast vascular structures, yet even these imperfect masks provide sufficiently strong 
supervisory signals to regularize the encoder and improve downstream resectability prediction. This demonstrates that although segmentation is an auxiliary task and not used at inference, the model does use the resulting latent anatomical features learned through the segmentation decoder; these features substantially improve the structure and clinical relevance of the imaging representations used by the classifier.

For the classification task, the proposed multimodal framework achieved strong and consistent performance across nested cross-validation folds, with an accuracy of 0.85, a macro-F1 score of 0.79, and an AUC of 0.86 (Table~\ref{tab:results}).

To assess generalization across institutions, we performed an external validation on an independent cohort from Kantonsspital Aarau (KSA Aarau). This dataset was not used during model development, hyperparameter tuning, or internal cross-validation. On this external cohort, the proposed method achieved an AUC of 0.86, a macro-F1 score of 0.81, and an accuracy of 0.87

(Table~\ref{tab:external_validation}).

Despite differences in patient characteristics and institutional acquisition settings, the model maintained performance comparable to the internal cross-validation results, demonstrating robustness to inter-institutional domain shift. These findings support the generalizability of the proposed multimodal framework beyond the originating centers.

Class-wise F1-score variability across folds is reported in Appendix Table~\ref{tab:per_class_f1}, showing stable performance despite the smaller sample size of the locally advanced class.

\begin{table}[h]
\centering
\caption{Classification results. Comparison of different methods on stratified nested 5-fold cross-validation (CV). 
Results are reported as mean $\pm$ standard deviation across folds.}
\label{tab:results}
\begin{tabular}{|l|c|c|c|}
\hline
\textbf{Method} & \textbf{AUC} & \textbf{Macro-F1} & \textbf{Accuracy} \\
\hline
\citep{Viviers2023} (Segmentation-Based) 
& 0.79 $\pm$ 0.04 & 0.74 $\pm$ 0.03 & 0.76 $\pm$ 0.04 \\
TAT (adapted from~\citep{dong2023improvedprognosticpredictionpancreatic})
& 0.83 $\pm$ 0.03 & 0.77 $\pm$ 0.03 & 0.81 $\pm$ 0.03 \\
\textbf{Ours (Multimodal Swin-UNETR)} 
& \textbf{0.86 $\pm$ 0.03} & \textbf{0.79 $\pm$ 0.02} & \textbf{0.85 $\pm$ 0.03} \\
\hline
\end{tabular}
\end{table}

\begin{table}[h]
\centering
\caption{External validation results on an independent cohort from Kantonsspital Aarau (KSA Aarau).
}
\label{tab:external_validation}
\begin{tabular}{|l|c|c|c|}
\hline
\textbf{Method} & \textbf{AUC} & \textbf{Macro-F1} & \textbf{Accuracy} \\
\hline
\citep{Viviers2023} (Segmentation-Based) 
& 0.82 & 0.77 & 0.81 \\
TAT (adapted from~\citep{dong2023improvedprognosticpredictionpancreatic})
& 0.84 & 0.79 & 0.84 \\
\textbf{Ours (Multimodal Swin-UNETR)} 
& \textbf{0.86} & \textbf{0.81} & \textbf{0.87} \\
\hline
\end{tabular}
\end{table}

\subsection{Ablation Studies: Feature Isolation, Knockout Robustness, and Single-Modality Training}

We conducted a series of ablation experiments to disentangle the contribution and robustness of each modality (Table~\ref{tab:combined_ablation}). First, in the feature isolation setting, the pretrained Swin-UNETR encoder and tabular MLP were frozen, and new classification heads were trained independently on their latent embeddings. This assessed the discriminative strength of each modality without re-training the fusion dynamics (FI block in Table~\ref{tab:combined_ablation}). 

Next, to evaluate resilience under partial input failure, a knockout analysis was performed in which one modality was replaced at inference by its mean embedding while the other remained intact. This quantified how strongly the fused classifier depends on each pathway and how well it tolerates missing or corrupted inputs (KO block).

Finally, single modality re-training was carried out by training CT only and tabular only models from scratch. These unimodal baselines revealed the inherent predictive capacity of each modality and highlighted the complementary nature of anatomical and clinical information in the multimodal model (SM block).

\begin{table}[t]
\centering
\caption{Combined ablation study. 
FI = Feature Isolation, KO = Knockout, SM = Single-Modality.}
\label{tab:combined_ablation}
\begin{tabular}{|l|l|c|c|c|}
\hline
\textbf{Type} & \textbf{Configuration} & \textbf{AUC} & \textbf{F1-Macro} & \textbf{Accuracy} \\
\hline

\multirow{2}{*}{FI}
& Imaging (frozen)       & 0.83$\pm$0.03 & 0.77$\pm$0.03 & 0.82$\pm$0.03 \\
& Tabular (frozen)       & 0.75$\pm$0.04 & 0.72$\pm$0.03 & 0.73$\pm$0.04 \\
\hline

\multirow{2}{*}{KO}
& Tabular → mean embedding  & 0.83$\pm$0.03 & 0.76$\pm$0.03 & 0.81$\pm$0.03 \\
& Imaging → mean embedding  & 0.77$\pm$0.04 & 0.70$\pm$0.04 & 0.73$\pm$0.04 \\
\hline

\multirow{2}{*}{SM}
& CT-only                & 0.82$\pm$0.03 & 0.76$\pm$0.03 & 0.78$\pm$0.03 \\
& Tabular-only           & 0.74$\pm$0.04 & 0.71$\pm$0.04 & 0.73$\pm$0.04 \\
\hline

\multicolumn{2}{|l|}{\textbf{Full multimodal (baseline)}} 
& \textbf{0.86$\pm$0.03} & \textbf{0.79$\pm$0.02} & \textbf{0.85$\pm$0.03} \\
\hline

\end{tabular}
\end{table}

\subsection{Comparisons with Existing Methods}

To contextualize our framework, we re-implemented and adapted two representative approaches:  
(i) the Texture-Aware Transformer (TAT) by Dong \textit{et al.}~\citep{dong2023improvedprognosticpredictionpancreatic}, originally developed for multiphase CT based survival prediction using texture-enhanced transformers and a neural distance module; and  
(ii) the segmentation-based assessment pipeline by Viviers \textit{et al.}~\citep{Viviers2023}, which computes handcrafted geometric features (e.g., tumor–vessel contact length, encasement, vessel involvement) from multiorgan segmentations to infer resectability according to the Dutch Pancreatic Cancer Group (DPCG) criteria.  
For comparability with our NCCN-based setting, we retained their geometric feature extraction but replaced the original DPCG rule based decision process with a supervised classifier trained to predict the three NCCN resectability categories.

Both baseline methods were adapted to match our single phase CT setting, incorporate structured clinical variables, and address the NCCN-based three class prediction task. Comparative results are summarized in Table~\ref{tab:results}.

TAT combines a texture encoder with two geometry focused components: a structural branch that processes local 3D patches around the shortest tumor–vessel distances, and a neural distance module that captures proximity between the tumor and major vessels (SMA, SMV, portal vein, celiac trunk). For our task, we adapted TAT to single phase CT, replaced its survival regression head with a softmax classifier, and retrained all geometry modules to learn NCCN style interactions directly from tumor–vessel surfaces. Training schedules and resolutions were aligned with our framework for fairness. As shown in Table~\ref{tab:results}, our multimodal anatomy guided model outperformed the adapted TAT, largely due to the added clinical information and segmentation-based anatomical supervision.

For the Viviers-based baseline, the geometric descriptors extracted from our 16-class segmentations were used directly as input to the NCCN classifier. While this produced a strong handcrafted geometry baseline, our multimodal framework surpassed it by leveraging jointly learned imaging and clinical representations that capture anatomical and contextual patterns not easily encoded through handcrafted rules.

\section{Conclusion}

We presented an end-to-end multimodal framework for automated assessment of PDAC resectability that integrates 3D CT with structured clinical data. By combining anatomy aware segmentation supervision with a multimodal fusion head, the model learns vessel informed and clinically aligned representations for NCCN-based resectability prediction.

The method achieved strong classification performance across resectability categories, and ablation studies showed that imaging and clinical features provide complementary information: CT captures detailed tumor–vascular morphology, while clinical variables contribute essential patient specific context. Compared to unimodal and segmentation only baselines, the multimodal design consistently improved predictive accuracy.

Beyond performance, the framework introduces several methodological advances for this domain. Segmentation guidance encourages the encoder to learn vessel aware anatomical structure while avoiding the need for segmentation masks at inference time. The adaptive multitask objective stabilizes training by shifting focus from anatomical learning to classification as segmentation improves. Furthermore, the model represents the first end-to-end integration of 3D vessel aware CT features with structured clinical covariates specifically tailored to NCCN defined resectability assessment.

Limitations include the moderate cohort size and the use of semi-automated segmentation masks for auxiliary supervision.

Although clinical variables in the training cohort required imputation, no imputation was necessary for the external validation dataset. The comparable performance observed across cohorts suggests that the applied imputation strategy did not introduce a substantial bias in model predictions; however, further validation on additional external datasets needs to be conducted.

Although the reference standard reflects MDT consensus, we did not perform a dedicated multi-reader study comparing individual clinician performance with the model under standardized reading conditions. Such a prospective reader study (including assessment of interobserver agreement and comparison to MDT consensus) is an important next step for clinical deployment.

Explainability analyses and feature importance will be essential to assess generalization and clinical robustness.

More expressive fusion operators (e.g., cross-attention) may become beneficial at larger cohort sizes; in our experiments, they did not improve performance and reduced training stability, so we opted for the simplest reliable fusion for this dataset.
Future work will involve multi-institutional evaluation, integration of additional imaging modalities such as multiphase CT or histopathology, and exploration of more expressive fusion mechanisms.
Overall, by jointly modeling anatomical detail and patient context, this framework offers a reproducible and clinically meaningful step toward objective, reliable decision support in pancreatic cancer surgery.

\clearpage  % Acknowledgements, references, and appendix do not count toward the page limit (if any)
% Acknowledgments---Will not appear in anonymized version

\bibliography{midl26_9}

@article{Siegel2023,
  title={Cancer statistics, 2023},
  author={Siegel, Rebecca L. and Miller, Kimberly D. and Jemal, Ahmedin},
  journal={CA: A Cancer Journal for Clinicians},
  year={2023}
}

@article{Katz2013,
  title={Borderline resectable pancreatic cancer: need for standardization and methods for optimal clinical trial design},
  author={Katz, Matthew H.G. and Pisters, Peter W.T. and Evans, D. et al.},
  journal={Annals of Surgical Oncology},
  year={2013}
}

@article{Giannone2021,
  title = {Resectability of Pancreatic Cancer Is in the Eye of the Observer: A Multicenter,  Blinded,  Prospective Assessment of Interobserver Agreement on NCCN Resectability Status Criteria},
  volume = {2},
  ISSN = {2691-3593},
  url = {http://dx.doi.org/10.1097/AS9.0000000000000087},
  DOI = {10.1097/as9.0000000000000087},
  number = {3},
  journal = {Annals of Surgery Open},
  publisher = {Ovid Technologies (Wolters Kluwer Health)},
  author={Giannone, Fabio and Capretti, Giovanni and Hilal, Mohammed Abu and Boggi, Ugo and Campra, Donata and Cappelli, Carla and Casadei, Riccardo and De Luca, Raffaele and Falconi, Massimo and Giannotti, Gabriele and others},
  year = {2021},
  month = aug,
  pages = {e087}
}

@misc{dong2023improvedprognosticpredictionpancreatic,
      title={Improved Prognostic Prediction of Pancreatic Cancer Using Multi-Phase CT by Integrating Neural Distance and Texture-Aware Transformer}, 
      author={Hexin Dong and Jiawen Yao and Yuxing Tang and Mingze Yuan and Yingda Xia and Jian Zhou and Hong Lu and Jingren Zhou and Bin Dong and Le Lu and Li Zhang and Zaiyi Liu and Yu Shi and Ling Zhang},
      year={2023},
      eprint={2308.00507},
      archivePrefix={arXiv},
      primaryClass={eess.IV},
      url={https://arxiv.org/abs/2308.00507}, 
}

@misc{cao2021swinunetunetlikepuretransformer,
      title={Swin-Unet: Unet-like Pure Transformer for Medical Image Segmentation}, 
      author={Hu Cao and others},
      year={2021},
      eprint={2105.05537},
      archivePrefix={arXiv},
      primaryClass={eess.IV},
      url={https://arxiv.org/abs/2105.05537}, 
}

@misc{cardoso2022monaiopensourceframeworkdeep,
      title={MONAI: An open-source framework for deep learning in healthcare}, 
      author={M. Jorge Cardoso and others},
      year={2022},
      eprint={2211.02701},
      archivePrefix={arXiv},
      primaryClass={cs.LG},
      url={https://arxiv.org/abs/2211.02701}, 
}

@misc{jiang2021fusionmedicalimagingelectronic,
      title={Fusion of medical imaging and electronic health records with attention and multi-head machanisms}, 
      author={Cheng Jiang and Yihao Chen and Jianbo Chang and Ming Feng and Renzhi Wang and Jianhua Yao},
      year={2021},
      eprint={2112.11710},
      archivePrefix={arXiv},
      primaryClass={cs.CV},
      url={https://arxiv.org/abs/2112.11710}, 
}

@article{Huang2016,
  title = {Radiomics Signature: A Potential Biomarker for the Prediction of Disease-Free Survival in Early-Stage (I or II) Non—Small Cell Lung Cancer},
  volume = {281},
  ISSN = {1527-1315},
  url = {http://dx.doi.org/10.1148/radiol.2016152234},
  DOI = {10.1148/radiol.2016152234},
  number = {3},
  journal = {Radiology},
  publisher = {Radiological Society of North America (RSNA)},
  author = {Huang,  Yanqi and others},
  year = {2016},
  month = dec,
  pages = {947–957}
}

@misc{saeed2022tmssendtoendtransformerbasedmultimodal,
      title={TMSS: An End-to-End Transformer-based Multimodal Network for Segmentation and Survival Prediction}, 
      author={Numan Saeed and Ikboljon Sobirov and Roba Al Majzoub and Mohammad Yaqub},
      year={2022},
      eprint={2209.05036},
      archivePrefix={arXiv},
      primaryClass={eess.IV},
      url={https://arxiv.org/abs/2209.05036}, 
}

@article{Sung2021,
  title={Global cancer statistics 2020: GLOBOCAN estimates of incidence and mortality worldwide for 36 cancers in 185 countries},
  author={Sung, Hyuna and Ferlay, Jacques and Siegel, Rebecca L and Laversanne, Mathieu and Soerjomataram, Isabelle and Jemal, Ahmedin and Bray, Freddie},
  journal={CA: A Cancer Journal for Clinicians},
  volume={71},
  number={3},
  pages={209--249},
  year={2021},
  publisher={Wiley Online Library},
  doi={10.3322/caac.21660}
}

@article{Bereska2024,
  title = {Artificial intelligence for assessment of vascular involvement and tumor resectability on CT in patients with pancreatic cancer},
  volume = {8},
  ISSN = {2509-9280},
  url = {http://dx.doi.org/10.1186/s41747-023-00419-9},
  DOI = {10.1186/s41747-023-00419-9},
  number = {1},
  journal = {European Radiology Experimental},
  publisher = {Springer Science and Business Media LLC},
  author = {Bereska,  Jacqueline I. and Janssen,  Boris V. and Nio,  C. Yung and Kop,  Marnix P. M. and Kazemier,  Geert and Busch,  Olivier R. and Struik,  Femke and Marquering,  Henk A. and Stoker,  Jaap and Besselink,  Marc G. and Verpalen,  Inez M.},
  year = {2024},
  month = feb 
}

@misc{viviers2023,
      title={Segmentation-based Assessment of Tumor-Vessel Involvement for Surgical Resectability Prediction of Pancreatic Ductal Adenocarcinoma}, 
      author={Christiaan Viviers and Mark Ramaekers and Amaan Valiuddin and Terese Hellström and Nick Tasios and John van der Ven and Igor Jacobs and Lotte Ewals and Joost Nederend and Peter de With and Misha Luyer and Fons van der Sommen},
      year={2023},
      eprint={2310.00639},
      archivePrefix={arXiv},
      primaryClass={eess.IV},
      url={https://arxiv.org/abs/2310.00639}, 
}

@misc{Ronneberger2015,
      title={U-Net: Convolutional Networks for Biomedical Image Segmentation}, 
      author={Olaf Ronneberger and Philipp Fischer and Thomas Brox},
      year={2015},
      eprint={1505.04597},
      archivePrefix={arXiv},
      primaryClass={cs.CV},
      url={https://arxiv.org/abs/1505.04597}, 
}

@article{Rawla2019,
  title={Epidemiology of pancreatic cancer: Global trends, etiology and risk factors},
  author={Rawla, Prashanth and Sunkara, Thandra and Gaduputi, Vinaya},
  journal={World Journal of Oncology},
  year={2019},
  volume={10},
  pages={10--27}
}

@article{Tempero2021,
  title={Pancreatic adenocarcinoma, version 2.2021, NCCN clinical practice guidelines in oncology},
  author={Tempero, Margaret A and others},
  journal={Journal of the National Comprehensive Cancer Network},
  year={2021},
  volume={19},
  number={4},
  pages={439--457}
}

@article{Neoptolemos2018,
  title={Therapeutic developments in pancreatic cancer: Current and future perspectives},
  author={Neoptolemos, John P and others},
  journal={Nature Reviews Gastroenterology \& Hepatology},
  year={2018},
  volume={15},
  pages={333--348}
}

@article{Bockhorn2014,
  title={Borderline resectable pancreatic cancer: a consensus statement by the International Study Group of Pancreatic Surgery (ISGPS)},
  author={Bockhorn, M. and others},
  journal={Surgery},
  year={2014},
  volume={155},
  number={6},
  pages={977--988}
}

@article{Isaji2018,
  title={International consensus on definition and criteria of borderline resectable pancreatic ductal adenocarcinoma 2017},
  author={Isaji, S. and others},
  journal={Pancreatology},
  year={2018},
  volume={18},
  pages={2--11}
}

@article{Callery2009,
  title={Pretreatment assessment of resectable and borderline resectable pancreatic cancer: expert consensus statement},
  author={Callery, M. P. and others},
  journal={Annals of Surgical Oncology},
  year={2009},
  volume={16},
  pages={1727--1733}
}

@article{Litjens2017,
  title     = {A survey on deep learning in medical image analysis},
  author    = {Geert Litjens and others},
  journal   = {Medical Image Analysis},
  year      = {2017},
  volume    = {42},
  pages     = {60--88},
  doi       = {10.1016/j.media.2017.07.005}
}

@article{Li,
  title={Medical image analysis using deep learning algorithms},
  author={Li, M. and others},
  journal={Front Public Health},
  year={2023}
}

@article{Zhou2020,
  title={Automatic segmentation of pancreas in abdominal CT images using deep convolutional neural network and self-supervised learning},
  author={Zhou, Y. and others},
  journal={Medical Physics},
  year={2020},
  volume={47},
  pages={5431--5442}
}

@article{Huang,
  title={Artificial intelligence in pancreatic cancer.},
  author={Huang, B. and Huang, H. and Zhang, S. and Zhang,  D. and  Shi, Q and  Liu,  J and Guo J.},
  journal={Theranostics},
  year={2022},
  volume={12},
  pages={6931-6954}
}

@article{Hartwig2013,
  title={Pancreatic cancer surgery in the new millennium: better prediction of outcome.},
  author={Hartwig, Wolfgang and others},
  journal={Annals of Surgery},
  year={2011},
  volume={254}
}

@article{Rahib2014,
  title={Projecting cancer incidence and deaths to 2030: the unexpected burden of thyroid, liver, and pancreas cancers in the United States.},
  author={Rahib, Lola and others},
  journal={Cancer Research},
  year={2014},
  volume={74},
  number={11},
  pages={2913--2921}
}

@article{Lim,
  title     = {Automated pancreas segmentation and volumetry using deep learning: a large N study},
  author    = {S. H. Lim and others},
  journal   = {European Radiology},
  year      = {2022},
  volume    = {32},
  number    = {6},
  pages     = {4563–4572},
  doi       = {10.1007/s00330-022-08653-w}
}

@article{Schuurmans2025_EndToEndPDAC,
  title   = { End-to-end prognostication in pancreatic cancer by multimodal deep learning: a retrospective, multicenter study},
  author  = {M. Schuurmans and others},
  journal = {European Radiology},
  year    = {2025},
  note    = {Multimodal AI for PDAC prognostic stratification}
}

@misc{ji2022amoslargescaleabdominalmultiorgan,
      title={AMOS: A Large-Scale Abdominal Multi-Organ Benchmark for Versatile Medical Image Segmentation}, 
      author={Yuanfeng Ji and Haotian Bai and Jie Yang and Chongjian Ge and Ye Zhu and Ruimao Zhang and Zhen Li and Lingyan Zhang and Wanling Ma and Xiang Wan and Ping Luo},
      year={2022},
      eprint={2206.08023},
      archivePrefix={arXiv},
      primaryClass={eess.IV},
      url={https://arxiv.org/abs/2206.08023}, 
}

@article{KORDNOORI2024101931,
title = {Deep multi-task learning structure for segmentation and classification of supratentorial brain tumors in MR images},
journal = {Interdisciplinary Neurosurgery},
volume = {36},
pages = {101931},
year = {2024},
issn = {2214-7519},
doi = {https://doi.org/10.1016/j.inat.2023.101931},
url = {https://www.sciencedirect.com/science/article/pii/S2214751923002141},
author = {Shirin Kordnoori and Maliheh Sabeti and Mohammad Hossein Shakoor and Ehsan Moradi}
}

\midlacknowledgments{This work was financially supported by the University of Basel through the Research Fund for Junior Researchers, which promotes the academic careers of outstanding junior researchers Sebastian M. Staubli was the recipient. In addition, Jo\"el L. Lavanchy was funded by the Swiss National Science Foundation (P5R5PM 21766), the Novartis Foundation for medical-biological Research (23C162) and by the Vontobel Foundation (0867/2024).}

\newpage

\appendix

\section{Ablation Studies}
\begin{table}[tbph]
\centering
\caption{Ablation of class weighting schemes for the nested cross-entropy loss 
(5-fold CV). The proposed smoothed weighting achieves the best balance 
between performance and stability.}
\label{tab:class_weight_ablation}
\begin{tabular}{|l|c|c|c|}
\hline
\textbf{Scheme} & \textbf{Weights} & \textbf{AUC} & \textbf{Macro-F1} \\
\hline
Unweighted & $(1,\,1,\,1)$ & 0.83$\pm$0.03 & 0.75$\pm$0.03 \\
Inverse-freq. & $(1,\,1.8,\,3.2)$ & 0.85$\pm$0.03 & 0.77$\pm$0.03 \\
\textbf{Smoothed (Ours)} & $\mathbf{(1,\,2,\,2.5)}$ & \textbf{0.86$\pm$0.03} & \textbf{0.79$\pm$0.02} \\
\hline
\end{tabular}
\end{table}

\begin{table}[tbph]
\centering
\caption{Ablation study comparing different weighting strategies between segmentation and classification 
losses (Equation~\ref{eq:total_loss}). Results were averaged across nested 5-fold cross-validation. 
The proposed Dice dependent adaptive weighting scheme achieves the best tradeoff between segmentation 
stability and classification accuracy.}
\label{tab:adaptive_ablation}
\begin{tabular}{|l|c|c|c|}
\hline
\textbf{Weighting Strategy} & \textbf{AUC} & \textbf{Macro-F1} & \textbf{Mean Tumor Dice} \\
\hline
Fixed weights $(w_\text{seg}=1.0,\, w_\text{cls}=1.0)$ 
& 0.84 $\pm$ 0.03 & 0.77 $\pm$ 0.03 & 0.49 $\pm$ 0.05 \\
Fixed weights $(w_\text{seg}=2.0,\, w_\text{cls}=0.5)$ 
& 0.83 $\pm$ 0.03 & 0.76 $\pm$ 0.03 & 0.52 $\pm$ 0.04 \\
Fixed weights $(w_\text{seg}=0.5,\, w_\text{cls}=2.0)$ 
& 0.80 $\pm$ 0.04 & 0.74 $\pm$ 0.04 & 0.42 $\pm$ 0.06 \\
\textbf{Adaptive (proposed)} 
& \textbf{0.86 $\pm$ 0.03} & \textbf{0.79 $\pm$ 0.02} & \textbf{0.56 $\pm$ 0.04} \\
\hline
\end{tabular}
\end{table}

\begin{table}[tbph]
\centering
\caption{Per-class F1-scores across stratified nested 5-fold cross-validation.
Results are reported as mean $\pm$ standard deviation across outer folds.
}
\label{tab:per_class_f1}
\begin{tabular}{|l|c|}
\hline
\textbf{NCCN Class} & \textbf{F1-score (mean $\pm$ SD)} \\
\hline
Upfront resectable      & $0.83 \pm 0.04$ \\
Borderline resectable  & $0.78 \pm 0.06$ \\
Locally advanced       & $0.70 \pm 0.07$ \\
\hline
\end{tabular}
\end{table}

\section{NCCN Geometry Analysis}
\FloatBarrier
\begin{figure}[tbph]
\centering
\includegraphics[width=0.78\linewidth]{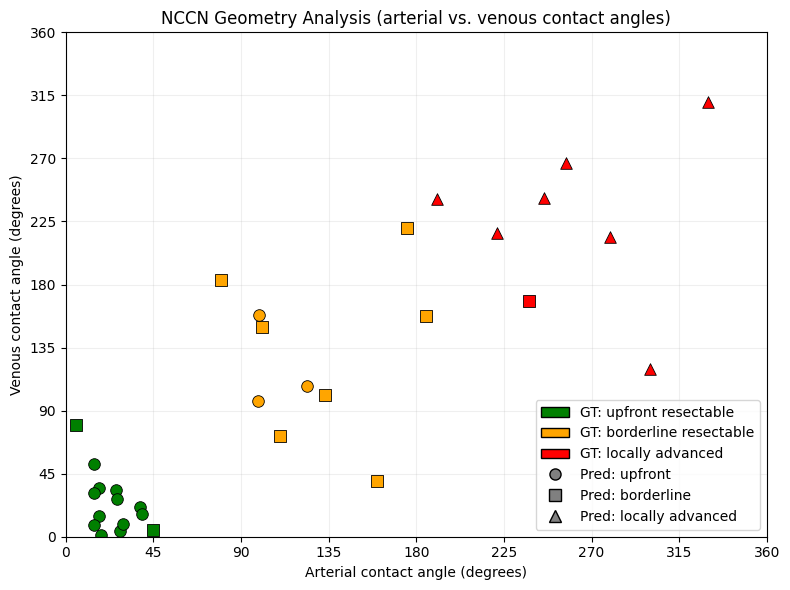}
\caption{
NCCN geometry analysis on the outer test folds. Arterial and venous tumor vessel contact angles were approximated from the model predicted segmentation masks and plotted against the ground truth NCCN resectability labels. Colors indicate the three NCCN classes (upfront 
resectable, borderline resectable, locally advanced), while marker shape denotes the model’s predicted class. Because the original dataset does not contain explicit geometric measurements and NCCN resectability additionally depends on factors beyond contact angles 
(e.g., vessel occlusion and reconstructability), a perfect one-to-one correspondence between regions and ground-truth labels is not expected. The plot serves as a sanity check showing that the learned representations exhibit NCCN consistent geometric trends.
}
\label{fig:nccn_scatter}
\end{figure}

\FloatBarrier
\section{Qualitative Examples}
\begin{figure}[tbph]
\centering
\includegraphics[width=0.9\linewidth]{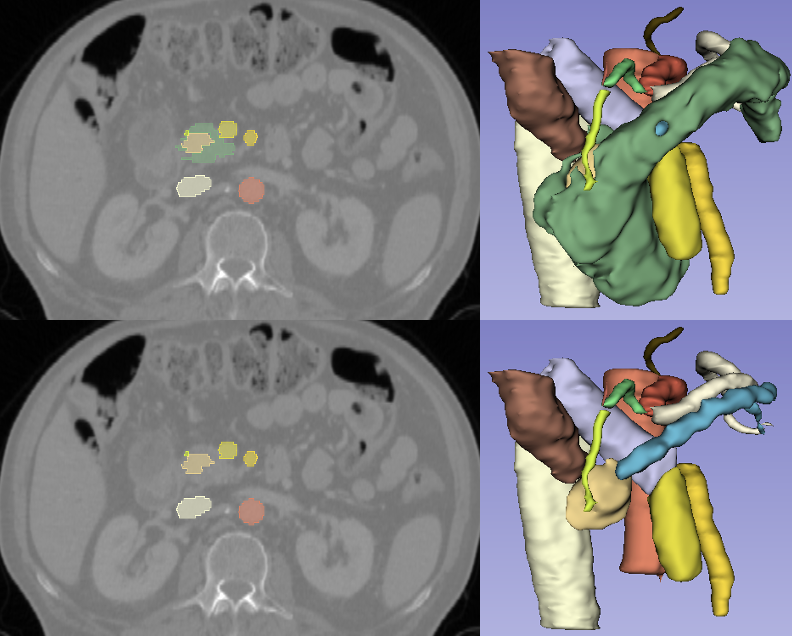}
\caption{
Qualitative examples of model predicted multiorgan and vascular segmentations.
Left: axial CT slices with overlayed predicted labels. 
Right: corresponding 3D renderings illustrating pancreas, tumor, and major vessels.
}
\label{fig:qualitative_segmentation}
\end{figure}

\end{document}